\documentclass[letterpaper]{article}
\usepackage[utf8]{inputenc}

\usepackage[margin=0.8in]{geometry}
\usepackage{fancyhdr}
\title{
    Comment on Stochastic Polyak Step-Size: \\
    Performance of ALI-G}
\author{
Leonard Berrada, Andrew Zisserman and M. Pawan Kumar \\
Department of Engineering Science, University of Oxford \\
\texttt{\{lberrada,az,pawan\}@robots.ox.ac.uk}
}
\date{\today}

\usepackage{amsmath,amsfonts,bm}

\def\eqref#1{equation~\ref{#1}}

\def\1{\bm{1}}

\def\vw{{\bm{w}}}

\DeclareMathAlphabet{\mathsfit}{\encodingdefault}{\sfdefault}{m}{sl}
\SetMathAlphabet{\mathsfit}{bold}{\encodingdefault}{\sfdefault}{bx}{n}

\DeclareMathOperator*{\argmin}{argmin}

\usepackage[utf8]{inputenc} 
\usepackage[T1]{fontenc}    
\usepackage{hyperref}       
\usepackage{url}            
\usepackage{booktabs}       
\usepackage{amsfonts}       
\usepackage{nicefrac}       
\usepackage{microtype}      

\usepackage{amsmath}
\usepackage{amsfonts}
\usepackage{amssymb}
\usepackage{mleftright}
\usepackage{xparse}

\usepackage{float}
\usepackage{algorithm}
\usepackage{graphicx}
\usepackage{subcaption}
\usepackage{dirtytalk}
\usepackage{multirow}
\usepackage{bbm}
\usepackage{bm}
\usepackage{mdframed}
\usepackage{wrapfig}

\usepackage{url}
\usepackage{booktabs}
\usepackage{amsfonts}
\usepackage{nicefrac}
\usepackage{microtype}
\usepackage{natbib}
\usepackage{apalike}
\usepackage{dsfont}
\usepackage{xcolor}
\usepackage{siunitx}
\usepackage{appendix}
\usepackage{xspace}
\usepackage{multirow}

\usepackage{tikz}
\usepackage{pgfplots}
\pgfplotsset{compat=1.12}
\usepackage{adjustbox}
\usepgfplotslibrary{groupplots}

\allowdisplaybreaks

\newcolumntype{R}[2]{%
    >{\adjustbox{angle=#1,lap=\width-(#2)}\bgroup}%
    l%
    <{\egroup}%
}

\setcitestyle{citesep={,}}  

\usepackage{float}
\usepackage{stmaryrd}
\usepackage{amsthm}

\usepackage{thmtools}
\usepackage{thmbox}
\usepackage{thm-restate}

\newmdenv[topline=false,rightline=false,bottomline=false,nobreak=false]{leftlinebox}

\NewDocumentCommand{\evalat}{sO{\big}mm}{%
  \IfBooleanTF{#1}
   {\mleft. #3 \mright|_{#4}}
   {#3#2|_{#4}}%
}

\newcommand{\wstar}{\mathbf{w_\star}}

\newcommand{\Z}{\mathcal{Z}}

\newsavebox{\leftbox}
\newsavebox{\rightbox}

\begin{document}

\maketitle

\begin{abstract}
    This is a short note on the performance of the ALI-G algorithm \citep{berrada2020training} as reported in \citep{loizou21a}.
    ALI-G \citep{berrada2020training} and SPS \citep{loizou21a} are both adaptations of the Polyak step-size to optimize machine learning models that can interpolate the training data.
    The main algorithmic differences are that (1) SPS employs a multiplicative constant in the denominator of the learning-rate while ALI-G uses an additive constant, and (2)
    SPS uses an iteration-dependent maximal learning-rate while ALI-G uses a constant one.
    There are also differences in the analysis provided by the two works, with less restrictive assumptions proposed in \citep{loizou21a}.
    In their experiments, \citep{loizou21a} did not use momentum for ALI-G (which is a standard part of the algorithm) or standard hyper-parameter tuning (for e.g. learning-rate and regularization). 
    Hence this note as a reference for the improved performance that ALI-G can obtain with well-chosen hyper-parameters.
    In particular, we show that when training a ResNet-34 on CIFAR-10 and CIFAR-100, the performance of ALI-G can reach  respectively 93.5\% (+6\%) and 76\% (+8\%) with a very small amount of tuning.
    Thus ALI-G remains a very competitive method for training interpolating neural networks.
\end{abstract}

\section{Context}

For context, we provide a summary on the Adaptive Learning-Rate for Interpolation with Gradients (ALI-G) algorithm. 
More details can be found in \citep{berrada2020training}.

\paragraph{Loss Function.}
We consider a supervised learning task where the model is parameterized by $\vw \in \mathbb{R}^p$.
The objective function is expressed as an expectation over $z \in \Z$, a random variable indexing the samples of the training set:
\begin{equation}
    f(\vw) \triangleq \mathbb{E}_{z \in \Z}[\ell_z(\vw)],
\end{equation}
where each $\ell_z$ is the loss function associated with the sample $z$ and is assumed to be non-negative: $\forall \vw \in \mathbb{R}^p, \ell_z(\vw) \geq 0$.

\paragraph{Regularization.}
We incorporate regularization (if any) as a constraint on the feasible domain: $\Omega = \big\{ \vw \in \mathbb{R}^p: \phi(\vw) \leq r \big\}$ for some value of $r$.
For example, for $\ell_2$ regularization: $\Omega = \left\{ \vw \in \mathbb{R}^p: \  \| \vw \|_2^2 \leq r \right\}$, for which the projection is given by a simple rescaling of $\vw$.

\paragraph{Problem Formulation.}
The learning task can be expressed as the problem $(\mathcal{P})$ of finding a feasible vector of parameters $\wstar \in \Omega$ that minimizes $f$:
\begin{equation} \label{eq:main_problem}
    \wstar \in \argmin\limits_{\vw \in \Omega} f(\vw).
\end{equation}

\paragraph{Interpolation Assumption.}
We assume that the problem at hand satisfies the interpolation assumption, i.e. we assume that there exists a solution $\wstar$ that simultaneously minimizes all individual loss functions:
\begin{equation}
    \forall z \in \Z, \: \ell_z(\wstar) = 0.
\label{eq:def_interpolation}
\end{equation}

\paragraph{ALI-G Update.}
The hyper-parameters of ALI-G are a maximal learning-rate $\eta$ (kept constant), and a small constant $\delta$ for numerical stability.
Then given a sample $z_t$ drawn at iteration $t$, the ALI-G update can be written as:
\begin{equation}
\vw_{t+1} = \Pi_{\Omega}\left(\vw_t - \min \left\{ \frac{\ell_{z_t}(\vw_t)}{\| \nabla \ell_{z_t}(\vw_t) \|^2 + \delta}, \eta \right\} \nabla \ell_{z_t}(\vw_t) \right).
\end{equation}

\paragraph{Remark on Hyper-Parameters.}
In \citep{loizou21a} the dependence of the hyper-parameters of ALI-G on the smoothness parameter is stated to be \say{limiting the method’s practical applicability}. This is not correct: in order to establish our theoretical convergence results we do indeed assume some dependence on the problem properties for our choices of $\eta$ (maximal learning-rate) and $\delta$ (denominator constant). However, in practical applications we simply set  $\delta$ to a low value (typically  $\delta=10^{-5}$) and treat it much like the $\epsilon$ numerical constant in Adam \citep{kingma15}.
Furthermore, in ALI-G only the maximal learning-rate $\eta$ is tuned, and it is then kept constant throughout training: this constitutes a major improvement over the most common practice of using a manually designed learning-rate schedule.

\paragraph{Remark on Momentum.}
No momentum is used for ALI-G in the experiments of \cite{loizou21a}, with the justification that ALI-G with momentum does not come with proofs of convergence.
While ALI-G with momentum is indeed not guaranteed to converge, it can still be used in practice, much like most optimization algorithms (including the Stochastic Polyak Step-Size) are applied to non-convex deep learning problems without formal guarantees of convergence.
Finally, we point out that the optimal choice of learning-rate depends on the use of momentum. Hence, to compensate for the deactivation of momentum for a method would call for careful tuning of the learning-rate.

\section{Empirical Performance on CIFAR Data Sets}

In \citep{loizou21a} ALI-G is employed without momentum and with a learning-rate set to 0.1, which was not selected for best performance. 
This set of experiments shows that the performance of ALI-G can be greatly improved with only a small amount of tuning.
We use the official code of \cite{loizou21a} at \url{https://github.com/IssamLaradji/sps} and run ALI-G on the task of learning a ResNet on CIFAR-10 and CIFAR-100.
We find that using momentum (with the common value of 0.9) and a maximal $\ell_2$ norm for regularization (with a value of 100) gives results significantly better than those reported in \cite{loizou21a}, as shown in Table \ref{tab:results}.

\begin{table}[H]
    \centering
    \begin{tabular}{llcccc}
    \toprule
        \multirow{2}{*}{Data Set} &\multirow{2}{*}{Method (Reference)} &\multicolumn{3}{c}{Hyper-Parameters Used} & Validation  \\ \cmidrule(lr){3-5}
        & &Learning-Rate &Momentum &Max $\ell_2$ norm & Accuracy \\ \midrule
        \multirow{2}{*}{CIFAR-10} & ALI-G \citep{loizou21a}  & 0.1 & 0 & $\infty$ & 87.5\%  \\
        & ALI-G (this note)  & 0.1 & 0.9 & 100 & {\bf 93.5\%} \\ \midrule
        \multirow{2}{*}{CIFAR-100} & ALI-G  \citep{loizou21a}  & 0.1 & 0 & $\infty$ & 68\% \\
        & ALI-G (this note)  & 0.1 & 0.9 & 100 & {\bf 76\%} \\
    \bottomrule
    \end{tabular}
    \caption{
        Performance of ALI-G with well-chosen hyper-parameters (this note), compared to the performance reported in \cite{loizou21a}.
        As can be observed, a small amount of tuning leads to large improvements (+6\% and +8\% on CIFAR-10 and CIFAR-100).
    }
    \label{tab:results}
\end{table}

\paragraph{Summary.}
In this note we reaffirm the practical applicability of ALI-G, and we demonstrate that with well-chosen hyper-parameters ALI-G can obtain state-of-the-art results for training interpolating neural networks without manually designed learning-rate schedules.

\bibliographystyle{apalike}
\bibliography{biblio}

\end{document}